\documentclass[runningheads]{llncs}

\usepackage{graphicx}

\usepackage{xcolor}
\usepackage{lineno,hyperref}
\usepackage{steinmetz}
\hypersetup{
	linkbordercolor = {0 1 0},
	citebordercolor = {0 0 1}
}
\usepackage{epstopdf}
\usepackage[caption=false]{subfig}
\usepackage{fancyhdr}
\usepackage{graphicx,color}
\usepackage[numbers,sort&compress,sectionbib]{natbib}
\usepackage{amsmath}
\usepackage{tikz}
\usetikzlibrary{matrix,shapes,arrows,positioning,chains}

\usepackage{multirow}
\usepackage{pbox}
\definecolor{cpb}{rgb}{0,1,0}

\usepackage{enumerate}
\usepackage{hhline}
\usepackage{setspace}

\begin{document}
\title{A Method to Generate Synthetically Warped Document Image}
%
%
\author{Arpan Garai\inst{1} \and Samit Biswas\inst{1}
\and Sekhar Mandal\inst{1} \and Bidyut. B. Chaudhuri\inst{2, 3}}
\authorrunning{A. Garai et al.}
%
\institute{Department of Computer Science and Technology, Indian Institute of Engineering Sciences and Technology, Shibpur, Hawrah, West Bengal, 711103, India
\email{arpangarai@gmail.com}, 
\email{\{samit,sekhar\}@cs.iiests.ac.in},\\
\and
Techno India University, Kolkata\\
\and
Computer Vision and Pattern Recognition Unit, Indian Statistical Institute, Kolkata, India\\
\email{bidyutbaranchaudhuri@gmail.com}
}

\maketitle              
\begin{abstract}
The digital camera captured document images may often be warped and distorted due to different camera angles or document surfaces. A robust technique is needed to solve this kind of distortion. The research on dewarping of the document suffers due to the limited availability of benchmark public dataset. In recent times, deep learning based approaches are used to solve the problems accurately. To train most of the deep neural networks a large number of document images is required and generating such a large volume of document images manually is difficult. In this paper, we propose a technique to generate a synthetic warped image from a flat-bedded scanned document image. It is done by calculating warping factors for each pixel position using two warping position parameters (WPP) and eight warping control parameters (WCP). These parameters can be specified as needed depending upon the desired warping. The results are compared with similar real captured images both qualitative and quantitative way. 

\keywords{Synthetic image generation \and Document image processing \and Dewarping.}
\end{abstract}

\section{Introduction}
\label{intro}
People nowadays prefer to use digital gadgets like cameras with mobile phones for capturing documents. These images are often distorted due to camera angles and/or non-planner document surface. Various forms of distortion in document image may arise due to these situations and warping is one of them. The performance of the OCR systems is not satisfactory when a highly warped document image is input to the OCR systems. A robust algorithm is needed to generate dewarped images from warped images. Recently, artificial neural networks like convolutional neural network (CNN)~\cite{MaCVPR18} and GAN based approaches~\cite{Dosovitskiy2016} are used for dewarping. To train such a network,  a large number of warped images are required. 

There are three publicly available  warped document images datasets. They are (i) `DFKI document image contest dataset' \cite{Shafait2007} and (ii) `IUPR dataset' \cite{Bukhari2012} (iii) `Data set by Ke Ma' \cite{MaCVPR18}. The `DFKI document image contest dataset' consists $102$ warped images and only ASCII text ground truth is available.
 There are  $100$ and $130$ images present in the  `IUPR dataset', and `Ke Ma dataset', respectively. Generally, such number of images is not enough to train deep neural networks to solve the dewarping problem. Capturing  a large number of images manually is a difficult task. Moreover, ground truth (images of document taken on a flat surface) is needed for supervised training. Generation of the ground truth for images from captured images is a non-trivial task. So, synthetic image generation is very necessary.

 Some techniques are already proposed to generate distortion in document images. Such methods can roughly be classified into three types \cite{Kieu2013}. They are (i) adding noise, \cite{Zhai2003} (ii) degrading characters, \cite{Kieu_c_d2013} and (iii) distorting the shape of document images\cite{Kieu2013}.   Kieu et. al. \cite{Kieu2013} proposed a mesh-based  semi-synthetic method to generate geometric distortion in the image. The mesh is generated using `the Kréon Aquilon laser 3D scanner'.

 In this work, a technique is proposed where synthetic warped images are generated from a flat-bed scanned document image. In this technique, warping factors are specified for each pixel position of the image. The warping factors are estimated using cubic spline interpolation. The positions of the knot points are generated by warping position parameters (WPP) and values of those knot points are calculated by warping control parameters (WCP). These parameters can be generated as necessary to create various kinds of warping. The curvy-ness of the surface of a document is estimated using Cylindrical Surface Model \cite{Cao2003}. The process is described in the Section- \ref{syn_img_gen}.
 
 To measure the similarity between ground truth image and dewarped image various metric has been used previously \cite{Kim2015}, \cite{Kil2017}. They are based on OCR, feature of text line etc. 
 There is no evaluation method where pixel wise comparison is done between ground truth image and dewarped image due to the difference of the resolution of ground truth image and real captured image. The proposed synthetic image generation can be applied to solve this problem and fulfil the gap. The proposed method takes a binary image as an input then it generates warped image  synthetically. Next, warped image is dewarped using a dewarping technique. Finally, the dewarped image is compared with the input image using pixel wise comparison measures like F-Measure \cite{Pratikakis2017}, pseudo F-Measure (Fps) \cite{Ntirogiannis2013} \cite{Pratikakis2017}, PSNR \cite{Pratikakis2017}, DRD \cite{Kot2004} \cite{Pratikakis2017}, Recall \cite{Pratikakis2017}, Precision \cite{Pratikakis2017} etc.

\section{Proposed Method}\label{syn_img_gen}
The convolutional neural network (CNN) can be used to find warping position parameters from the warped document image.
A huge number of documents is required to train a CNN and capturing it is difficult to capture large number of  document images. 

\begin{figure}[h]
	\begin{center}
		$\begin{array}{@{\hspace{1pt}}c@{\hspace{1pt}}c@{\hspace{1pt}}c}
		\fbox{\includegraphics[height=0.87in]{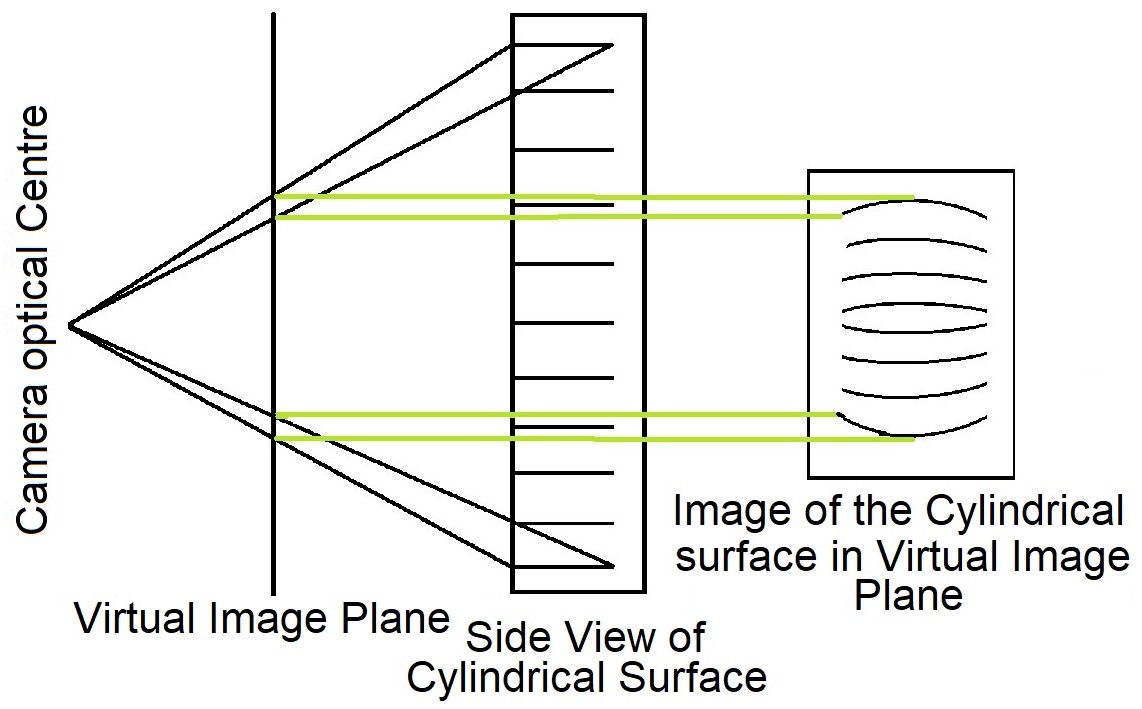}}&
		\fbox{\includegraphics[height=0.87in]{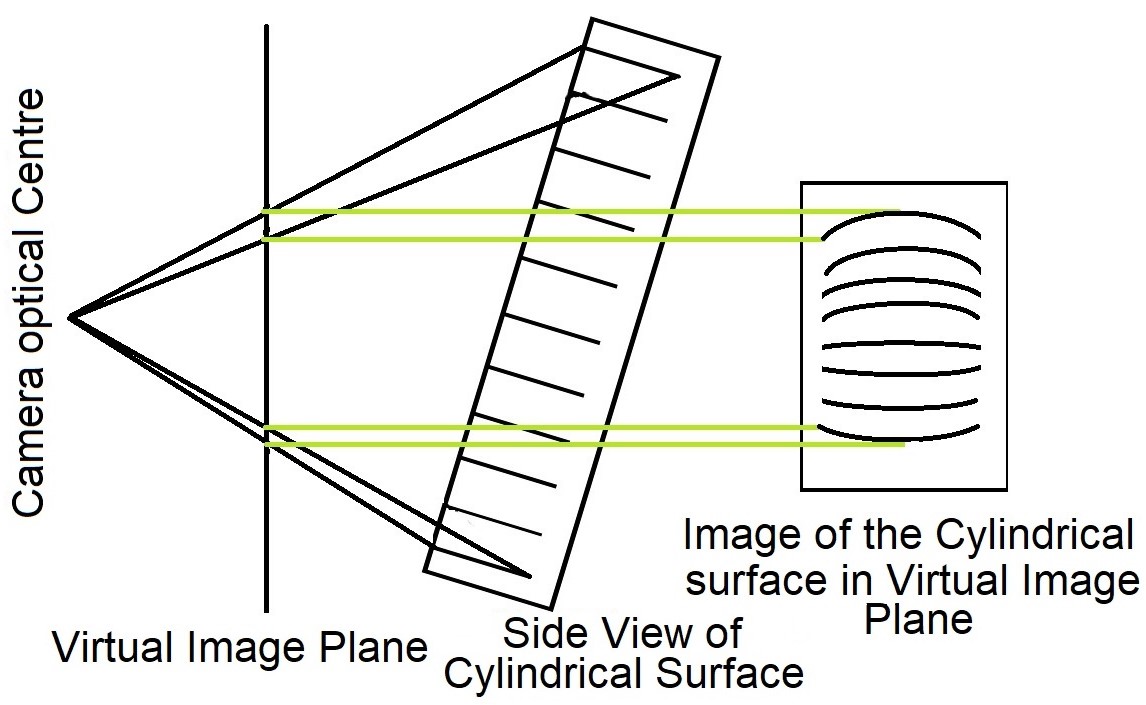}} &
		\fbox{\includegraphics[height=0.87in]{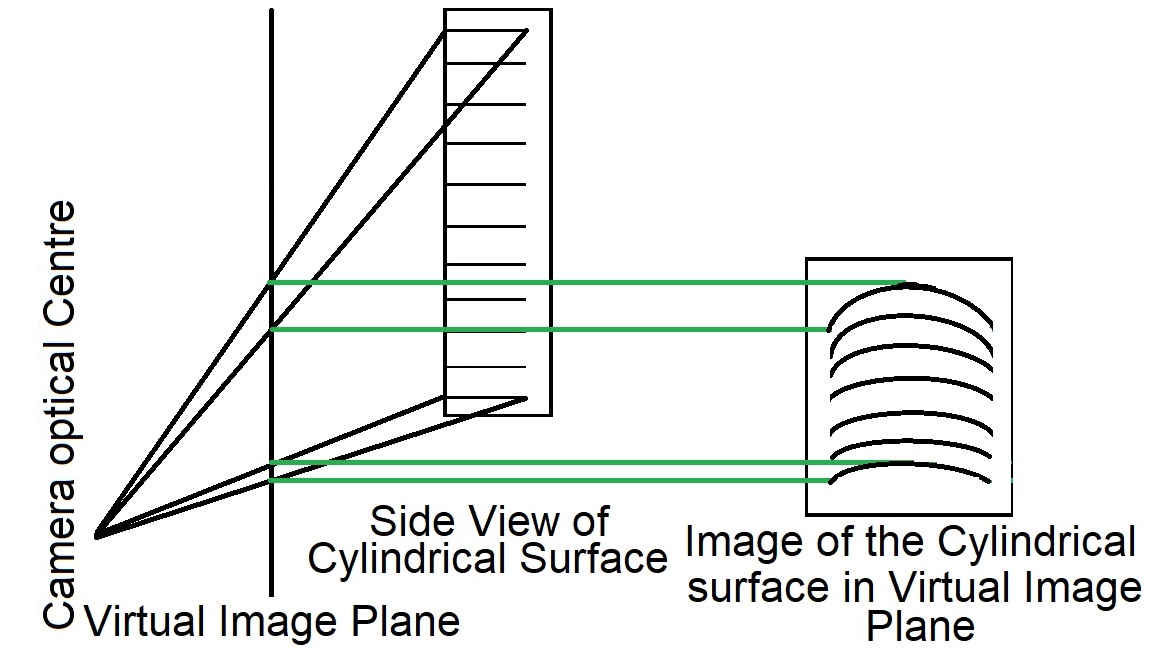}}\\
		\small \textnormal{(a)} & \small \textnormal{(b)} & \small \textnormal{(c)}\\
		\end{array} $  
		\caption{Example of warped documents for different reasons (a) Distance difference, (b) Foreshortening difference with camera at the middle, (c) Foreshortening difference with camera at the bottom. }
		\label{dist_diff}
	\end{center}
\vspace{-0.5cm}
\end{figure}

Hence, we generate synthetically the warped document images. Here, we developed a technique to generate several warped document images from a single captured image. We use  cylindrical surface model (CSM) proposed in \cite{Cao2003} to generate these images. According to CSM, an image with a cylindrical surface gets warped due to two major reasons. The first reason is the distance variation from the camera from different regions of surface (e.g the curvature of the surface) as shown in Fig. \ref{dist_diff}(a, c). The other reason is the foreshortening differences that occurs if the surface is non-perpendicularly oriented with the optical axis as shown in Fig. \ref{dist_diff}(b). Here, we  assume  that the surface does not suffer from foreshortening differences.

\begin{figure}[h]
	\begin{center}
		$\begin{array}{@{\hspace{1pt}}c}
		\fbox{\includegraphics[width=3.25in]{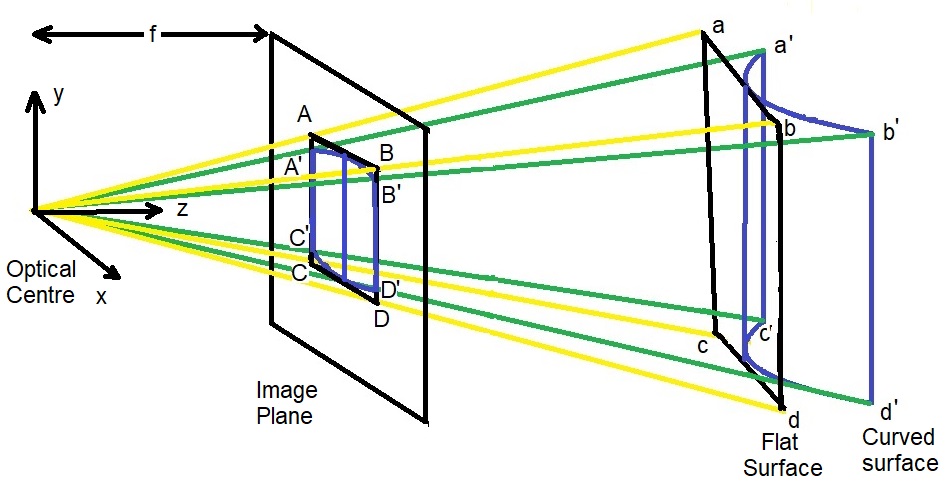}}\\
		\end{array} $  
		\caption{Model for warping generation.}
		\label{scn_mdl}
	\end{center}
\end{figure}
\vspace{-1cm}

Now, consider the Fig. \ref{scn_mdl}. Let $abcd$ be a document having  flat surface. Its  projection on the image plane is denoted by $ABCD$. Let $p(x,y,z)$ be a point on the surface $abcd$. The z-value for all the points on $abcd$ is constant which is denoted by D(0). The projection of the point $p$ in the image plane is represented by $P(X,Y)$. The values of $X$ and $Y$ can be obtained using the equations: $X = \frac{f}{D(0)}x \text{ , } Y = \frac{f}{D(0)}y$ \cite{Cao2003}. Here $f$ is the focal distance.

Consider another document ($a^\prime b^\prime c^\prime d^\prime$) having  curved surface and its  projection in the image plane is represented by $A^\prime B^\prime C^\prime D^\prime$. Let us consider a  point $p^\prime(x^\prime,y^\prime,z^\prime)$ on the surface $a^\prime b^\prime c^\prime d^\prime$ and its projection on the image plane is denoted by $P^\prime(X^\prime,Y^\prime)$. Note that the z value of a point in $a^\prime b^\prime c^\prime d^\prime$ is a function of x coordinate and denoted by $D(x)$. We can obtain the  $X^\prime$ and $Y^\prime$ using the equations:
$X^\prime = \frac{f}{D(x)}x^\prime \text{ , } Y^\prime = \frac{f}{D(x)}y^\prime$.

Consider two points $p_1(x_1,y_1,D(0))$ and $p^\prime_1(x_1,y_1,D(x))$ on the flat surface and the curved surface, respectively. The x and y values of aforesaid point are same and the difference of the corresponding $y$-values in the image plane is obtained as: $
(Y_1' - Y_1) = \frac{f.y_1}{D(x)} -\frac{f.y_1}{D(0)}
= fy_1\frac{D(0)-D(x)}{D(x).D(0)}
= fy_1\frac{D(0)-D(x)}{D(0)[(D(x)-D(0))+D(0)]}
$.

While capturing a page of a book/ document on lamp-post, notice-board etc., the value of $(D(x)-D(0))$~\cite{Cao2003} generally lies between $1$ cm to $3$ cm whereas $D(0)$ can have values $50$ cm to $100$ cm.
Hence, the value of [$D(x)-D(0)$] can be  neglected comparing to the value of $D(0)$.

\begin{equation}
\label{eq_d}
(Y_1' - Y_1) \approx fy_1\frac{D(0)-D(x)}{D(0)^2} = fy_1\frac{d}{D(0)^2}
= \big(\frac{fy_1}{D(0)}\big)\big(\frac{d}{D(0)}\big)
= Y_1\frac{d}{D(0)}
\end{equation}

Where $D(0) - D(x)$ = $d$ which is the variation of the z-coordinates in the object plane. So, it is clear from Eq. \ref{eq_d} that the translation of the point $P^\prime (X^\prime, Y^\prime)$ in $y$-direction depends on $d$. We can generate different warped images using different values of $d$. Here, $d$ is referred to as {\em warping factor}. The value of $d$ changes pixel to pixel.

In our experiment, we distribute the warping factor in the image plane considering the surface of a book. The warping factor is row-wise distributed. The value of $d$ in each cell within a row is determined using smooth cubic spline interpolation function. Five knot points are used for interpolation and the position of each knot point is determined by two warping parameters ($P_1$ and $P_2$). The amount of warping at each knot point is controlled by another eight parameters  ($P_3$, $\dots$, $P_{10}$). The detailed descriptions of the aforesaid parameters are described in the next section.

\subsection{Warping Position Parameters}
\begin{figure}[h]
	\begin{center}
		$\begin{array}{@{\hspace{10pt}}c}
		\fbox{\includegraphics[height=1.5in]{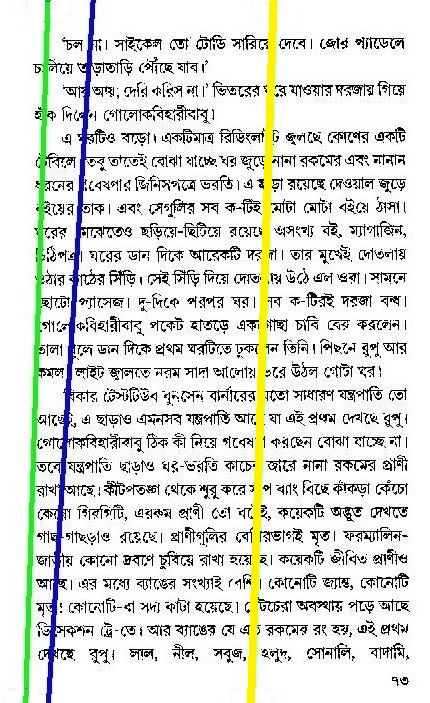}}\\
		\end{array}$
		\caption{Positions of $2^{nd}$(green), $3^{rd}$(blue) and $4^{th}$(yellow) knot points}
		\label{gp_234}
	\end{center}
\end{figure}

The parameters $P_1$ and $P_2$ are used to select the position of the document warping. The value of $P_1$ is multiplied by the width of the first row of the image and take its nearest integer to get the position of the cell (say $Z_1$) where there will be no distortion due to warping. Similarly, $P_2$ determines the position of the column ($Z_2$) in the bottom row where distortion will not  take place due to warping. The parameters $P_1$ and $P_2$ take any one of the of following real numbers ($0.1, 0.2, 0.3,\ldots, 0.9$). Using $P_1$ and $P_2$ we select the position of the third knot point. The position of second knot points are along the nearest point the middle of straight line $Z_1-Z_2$ (blue line at Fig. \ref{gp_234}) and left boundary (green line at Fig. \ref{gp_234}). The position of forth knot points are along the nearest point the middle of straight line of $Z_1-Z_2$ and right boundary (yellow line at Fig. \ref{gp_234}). Taking different values of $P_1$ and $P_2$, we can change the position of knot points and hence, we can generate different warped documents from a single document. 

\vspace{-1cm}

\subsection{Warping Control Parameters}
Smoothing cubic spline interpolation function is used to warp the document. Five knots are used for this interpolation function in each row of the image. The value of the leftmost and rightmost knots of the first row of the image is denoted by  $P_3$ and $P_6$, respectively. Similarly, $P_7$ and $P_{10}$ represent the values of the leftmost and rightmost knots of the last row of the image.
Let $D$ is the length of a diagonal of the image under test then, the value of this parameter is $k\times D$. In our experiment, $k$ varies within the range $0.04-0.6$ with a step size of $0.01$.

Similarly, $P_4$ and $P_5$ specify the values of $2^{nd}$ and $4^{th}$ knot point of the top-most row and $P_8$ and $P_9$ specify the values of $2^{nd}$ and $4^{th}$ knots of the bottom-most row, respectively. In our experiment, we have specified $P_4 = 0.5\times P_3$, $P_5 = 0.5\times P_6$, $P_8 = 0.5\times P_7$ and $P_9 = 0.5\times P_{10}$.

\begin{figure}[h]
	\begin{center}
		$\begin{array}{@{\hspace{2pt}}c@{\hspace{2pt}}c}
		\fbox{\includegraphics[width=2.33in]{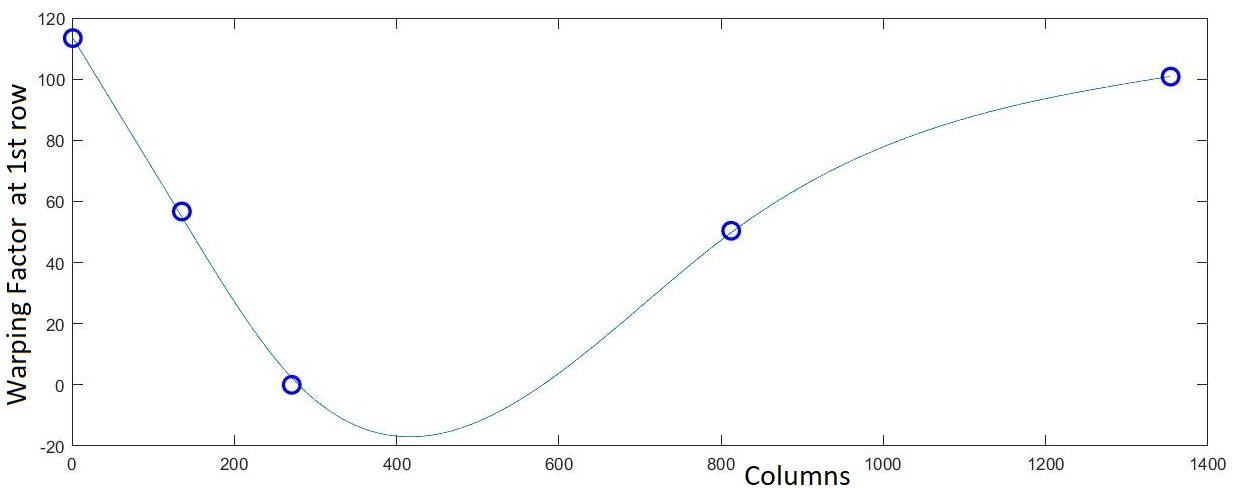}}
		&
		\fbox{\includegraphics[width=2.33in]{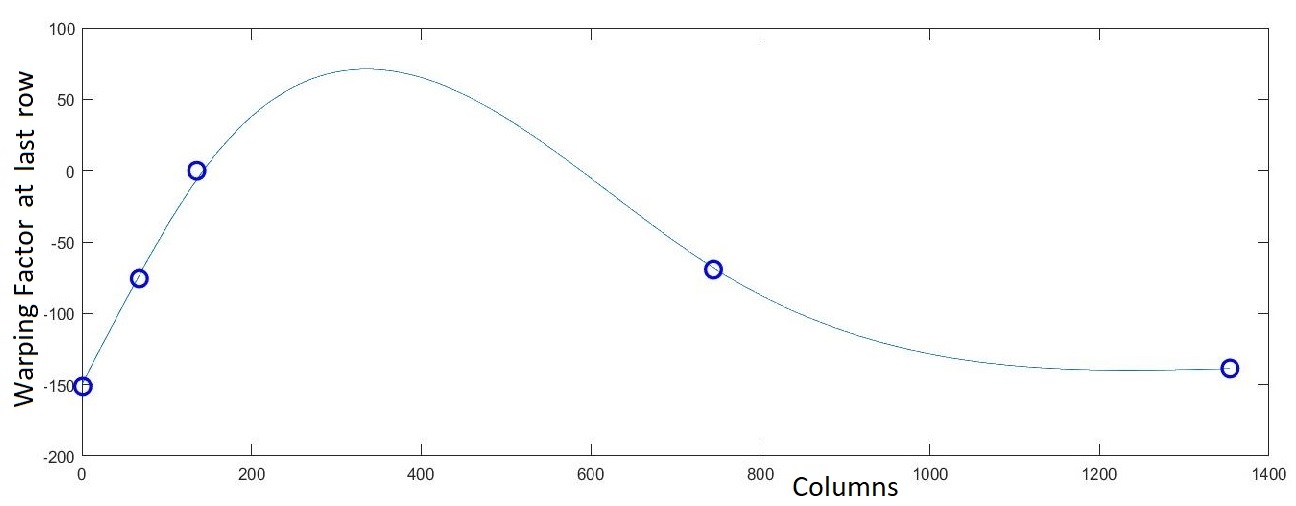}}\\
		\small \textnormal{(a)} &\small \textnormal{(b)}\\
		\end{array}$
		$\begin{array}{@{\hspace{2pt}}c}
		\fbox{\includegraphics[width=3.3in]{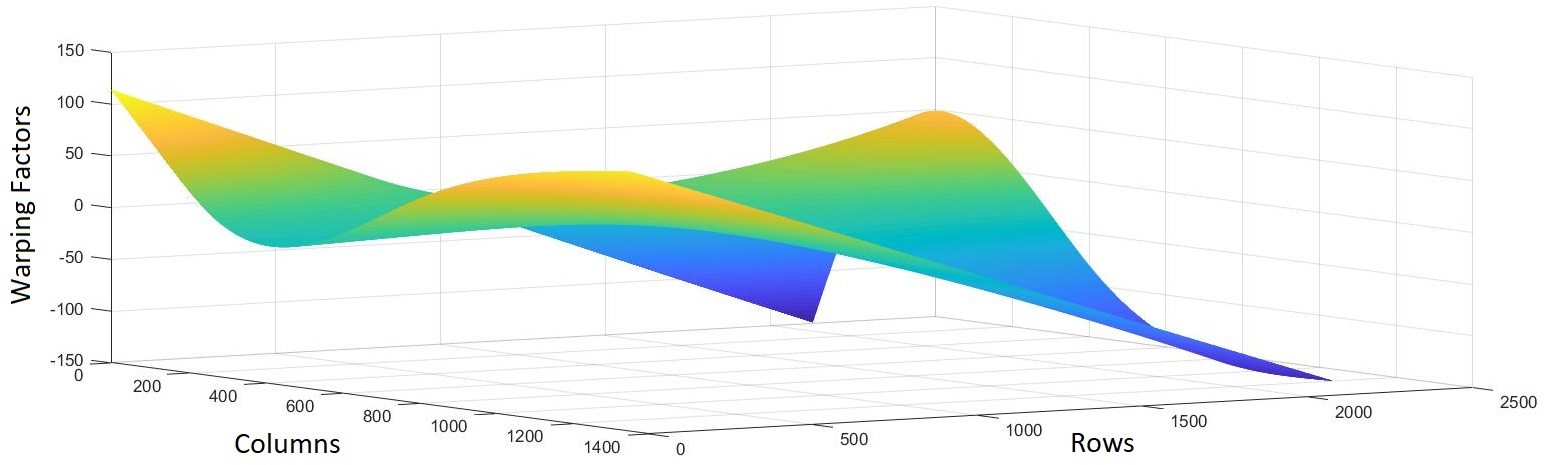}}\\
		\small \textnormal{(c)}\\
		\end{array}$
		\caption{(a) Warping Factors for each pixel the top-most row; (b) Warping Factors for each pixel the bottom-most row; (c) Three dimensional plotting of the warping factors.}
		\label{wrp_1_2}
	\end{center}
\end{figure}

Consider the values of $1^{st}$, $2^{nd}$, $4^{th}$ and $5^{th}$ knots of the $i^{th}$ row are $G^i_1$, $G^i_2$, $G^i_3$ and $G^i_4$, receptively. They are determined using the following equations. Here, $R$ is total number of rows present in the image. 

$$
G^i_1 = \frac{P_7-P_3}{R-1}\times (i-1) + P_3
\text{ , }
G^i_2 = \frac{P_8-P_4}{R-1}\times (i-1) + P_4
$$
$$
G^i_4 = \frac{P_9-P_5}{R-1}\times (i-1) + P_5
\text{ , }
G^i_5 = \frac{P_{10}-P_6}{R-1}\times (i-1) + P_6
$$

The third knot of the interpolation function is on line passing through the points $Z_1$ and $Z_2$ and hence, its value is zero.

\subsection{Calculating the warping factors}
Using the values of five knots at each row we interpolate the values of the other points of the row. We have applied smoothing spline regression to interpolate the values. An example of the warping factors at the top-most and bottom-most row is shown in Fig. \ref{wrp_1_2}(a) and \ref{wrp_1_2}(b) respectively. At each location of the image we have calculated a warping factor and they are stored in a 2-D array $F$. The of 3-D representation of $F$ is shown in Fig. \ref{wrp_1_2}(c).

\begin{figure}[h]
	\begin{center}
		$\begin{array}{@{\hspace{10pt}}c@{\hspace{10pt}}c}
		\fbox{\includegraphics[height=2in]{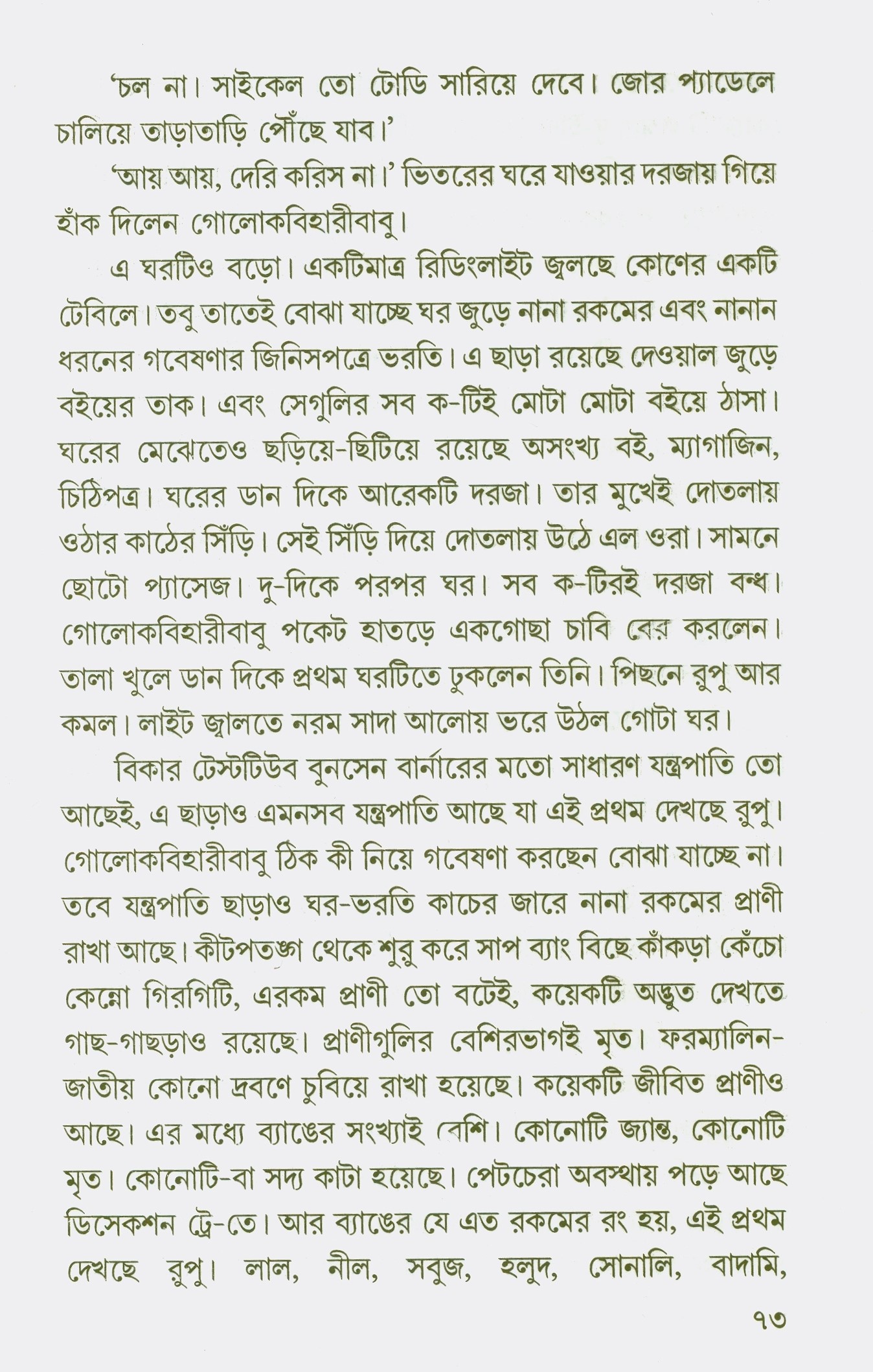}}& 
		\fbox{\includegraphics[height=2in]{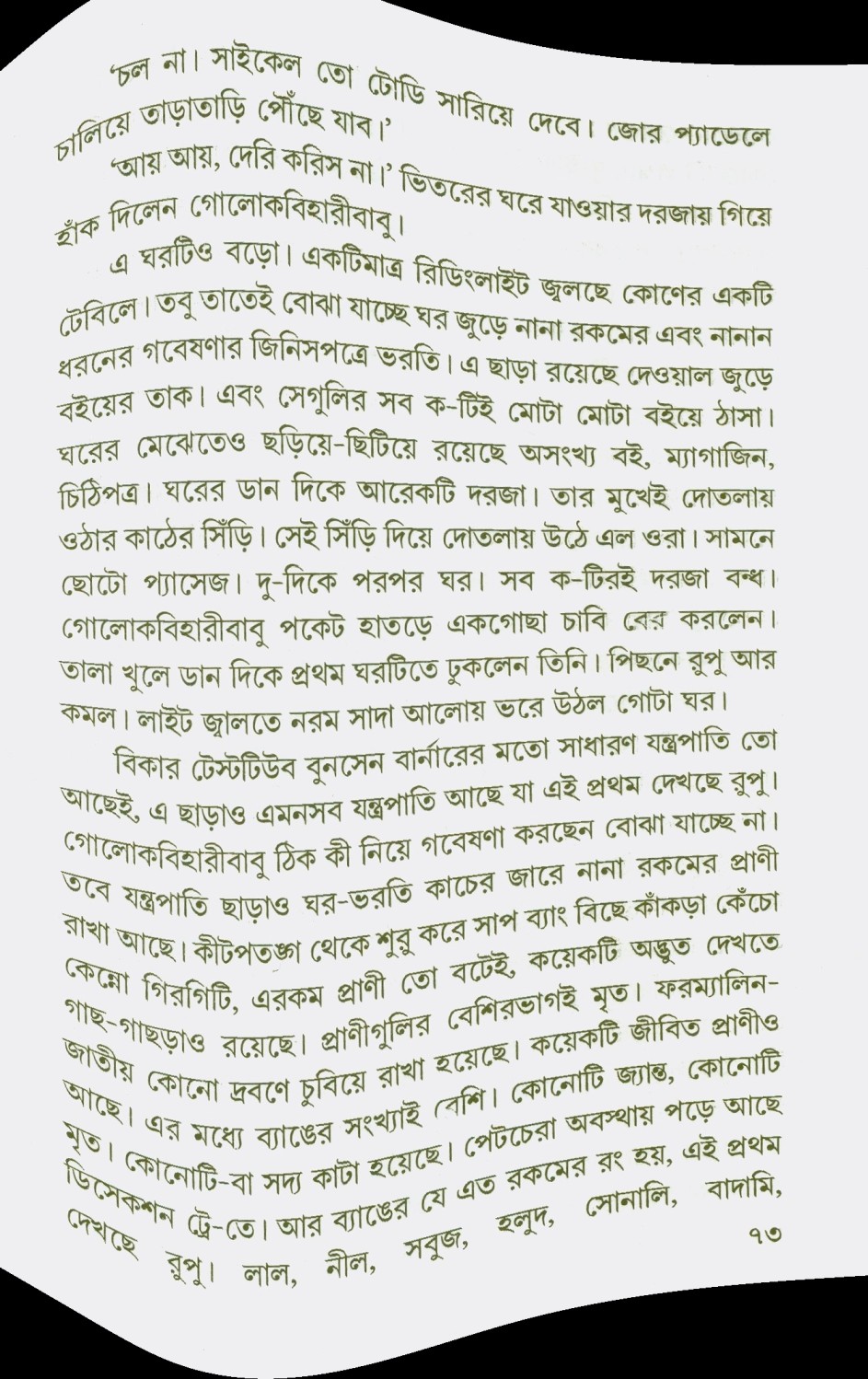}}\\
		\small \textnormal{(a)} & \small \textnormal{(b)}\\
		\end{array} $ 
		\caption{(a) Scanned image (b) Synthetic warped image} 
		\label{eg_wrp}
	\end{center}
\end{figure}

Finally, each pixel is translated using the respective warping factors. Here,  nearest neighbour interpolation is used during the process of translation. An example of scanned image and its corresponding synthetic warped image is shown in Fig. \ref{eg_wrp}. Here, in Fig. \ref{wrp_1_2}, and \ref{eg_wrp}, the value for $P_1$ and $P_2$ are $0.2$ and  $0.1$, respectively. The diagonal of the image is $D = 2521$ pixel. The values of warping control parameters are:$P_3 = 0.045\times D = 113$, $P_6 = 0.04\times D = 101$, $P_7 = -0.06\times D = -151$, $P_{10} = -0.055\times D = -139$, $P_4 = 113/2 = 56.5$, $P_5 = 101/2 = 50.5$, $P_8 = -151/2 = -75.5$ and $P_9 = -139/2 = -69.5$. 

\begin{figure*}[h!]
	\begin{center}
		$\begin{array}{@{\hspace{3pt}}c@{\hspace{3pt}}c@{\hspace{3pt}}c@{\hspace{3pt}}c}
		\fbox{\includegraphics[height=1.4in]{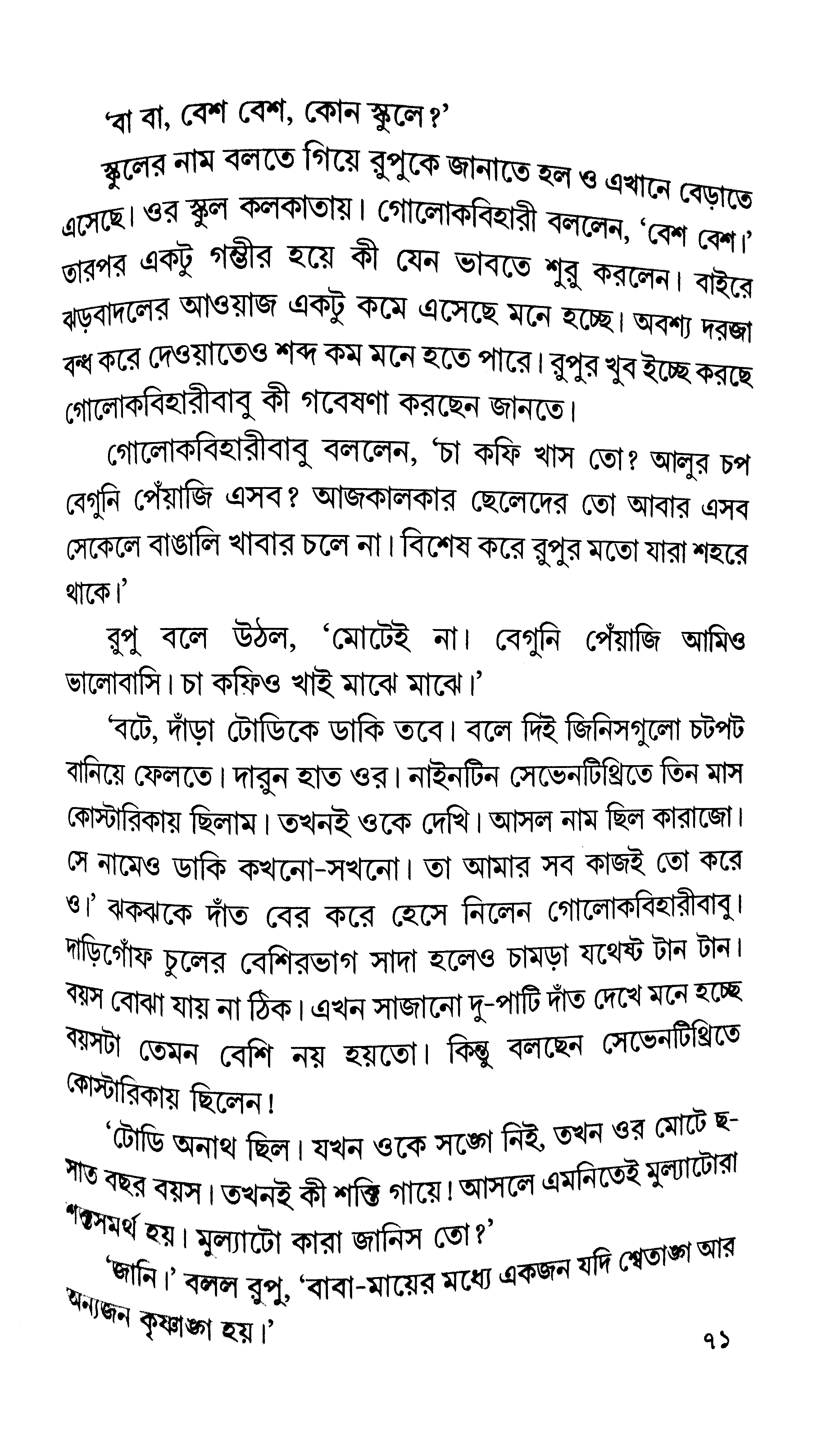}}&
		\fbox{\includegraphics[height=1.4in]{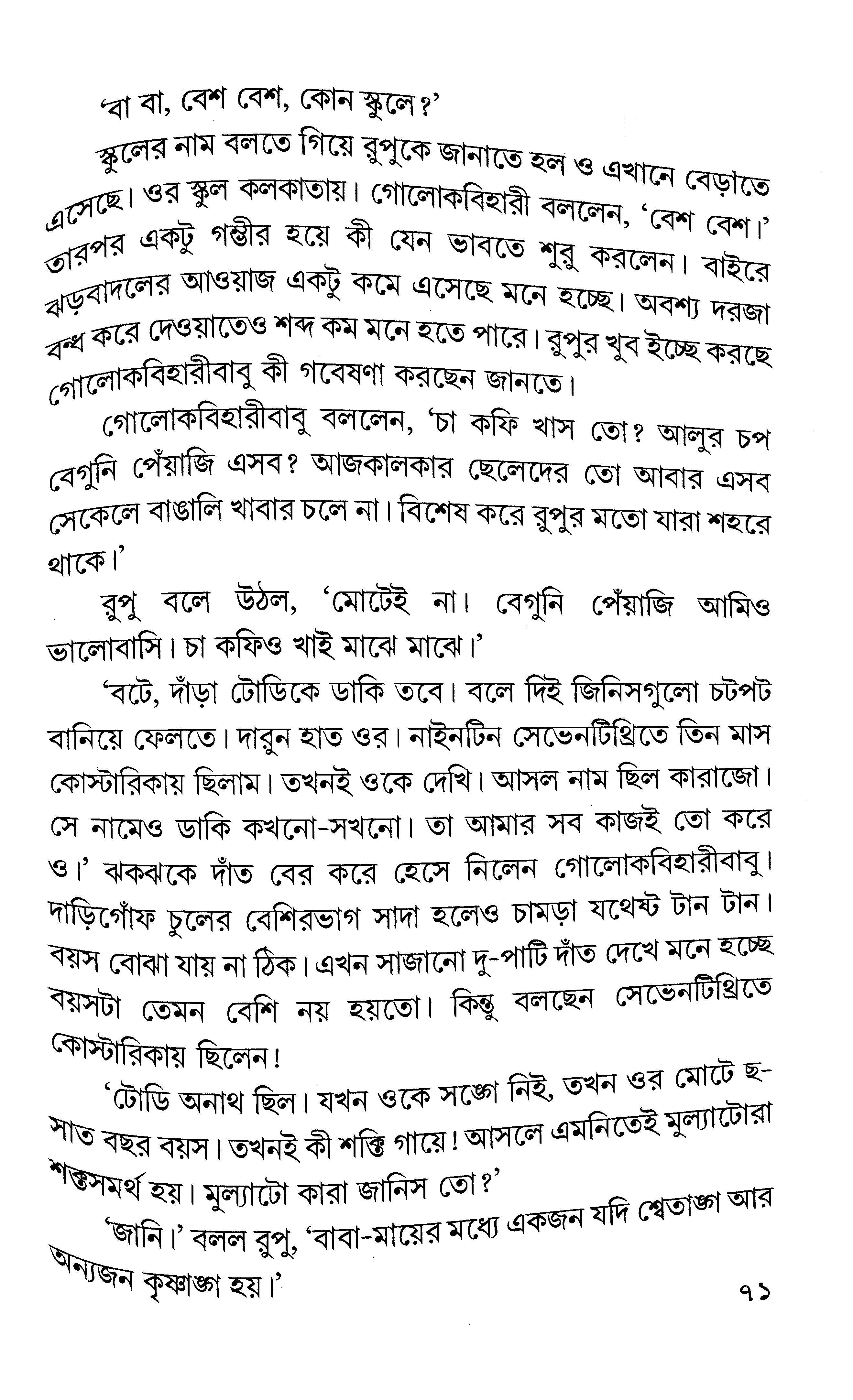}}&
		\fbox{\includegraphics[height=1.4in]{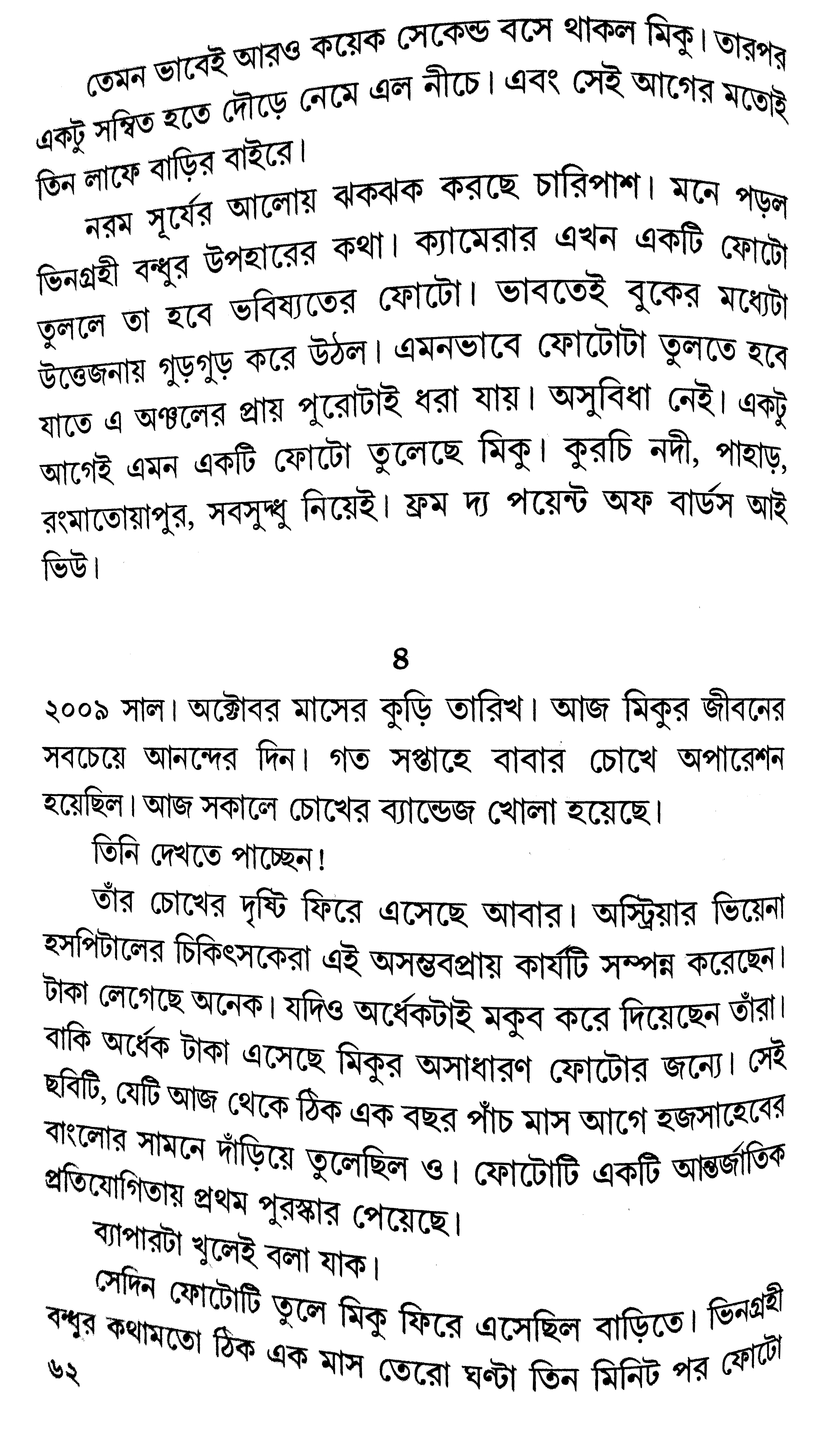}}&
		\fbox{\includegraphics[height=1.4in]{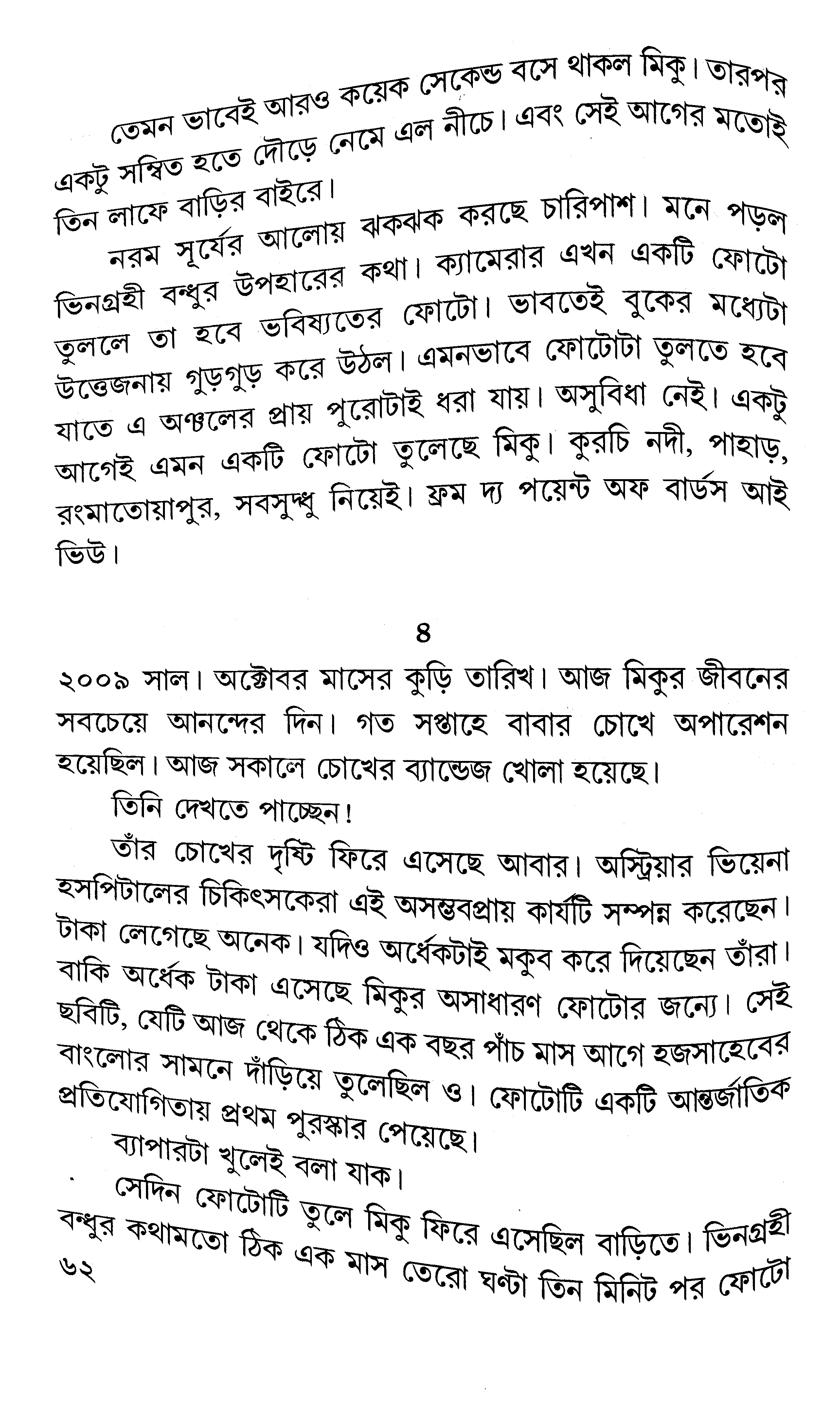}}\\
		\small \textnormal{(a)} & \small \textnormal{(b)}& \small \textnormal{(c)} & \small \textnormal{(d)}\\
		\fbox{\includegraphics[height=1.4in]{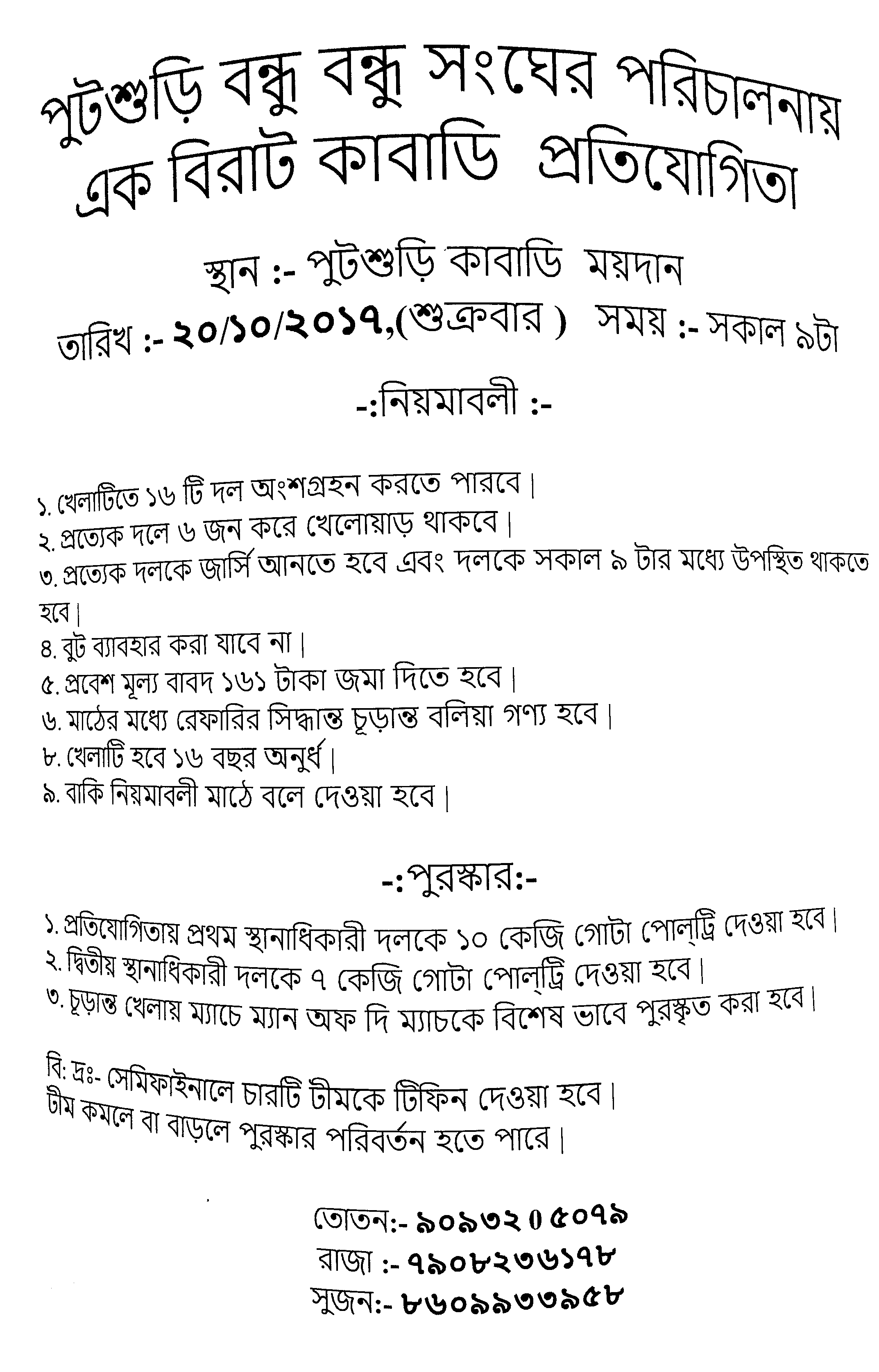}}&
		\fbox{\includegraphics[height=1.4in]{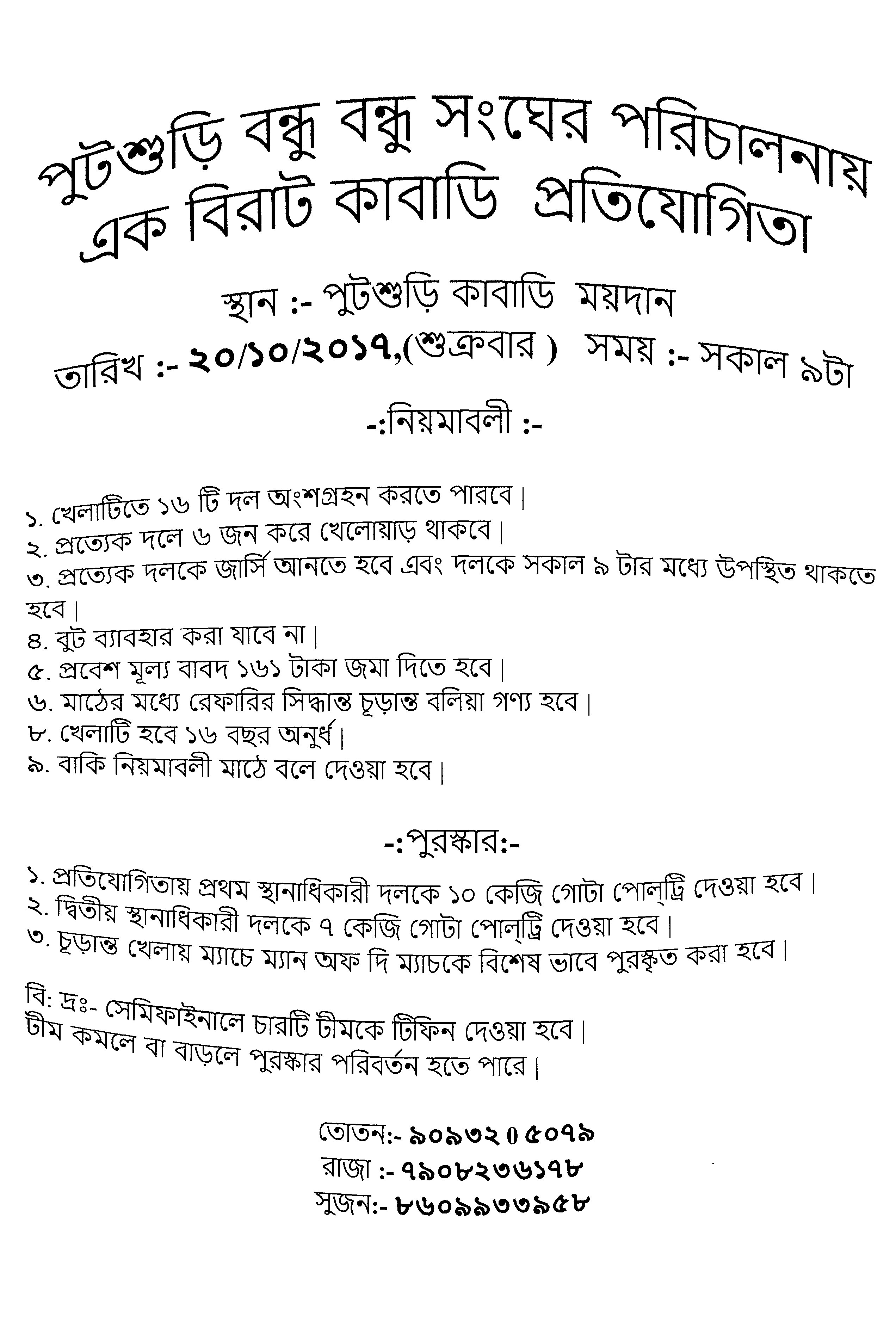}}&
		\fbox{\includegraphics[height=1.4in]{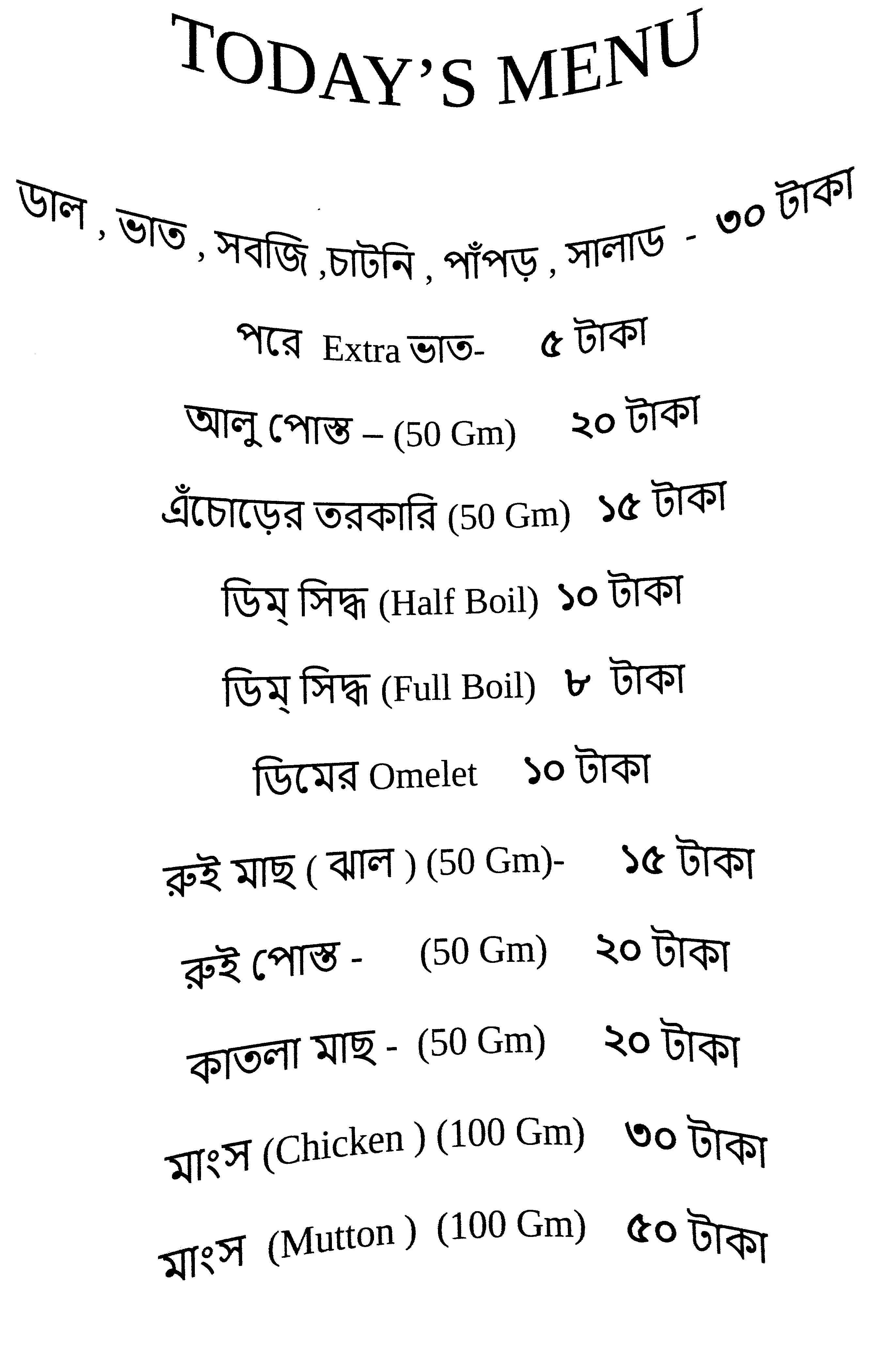}}&
		\fbox{\includegraphics[height=1.4in]{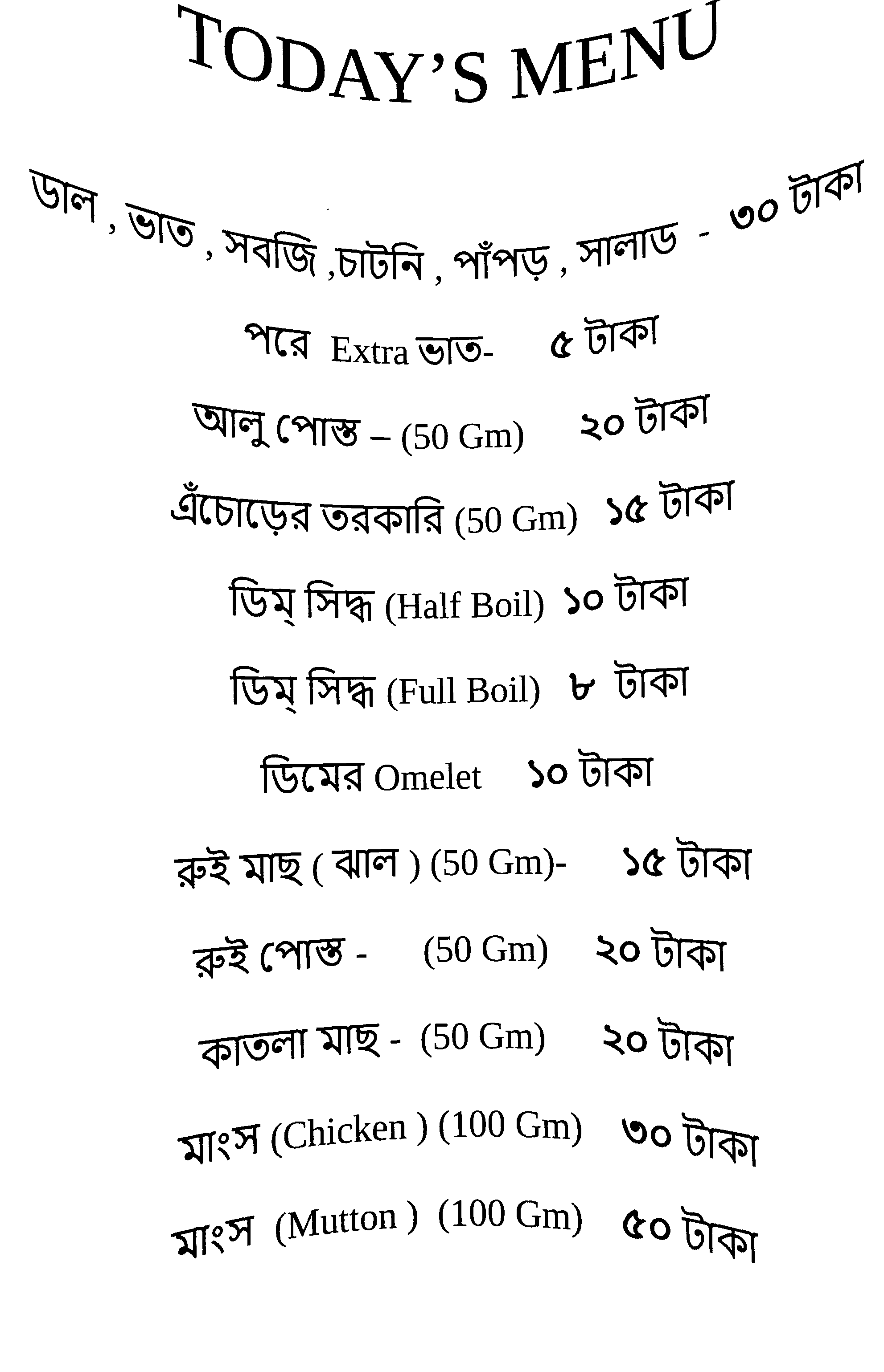}}\\
	 	\small \textnormal{(e)}& \small \textnormal{(f)}& \small \textnormal{(g)}& \small \textnormal{(h)}\\
		\end{array} $ 
		\caption{Visual comparison: Book page image undistorted part at left: (a) Pre-processed real captured warped image, (b) Synthetic warped image; Book page image undistorted part at right: (c) Pre-processed real captured warped image, (d) Synthetic warped image; 
			Document pasted on a cylindrical lamp-post: (e) Pre-processed real captured warped image, (f) Synthetic warped image;	Document hanging from notice-board: (g) Pre-processed real captured warped image, (h) Synthetic warped image.\vspace{-1cm}} 
		\label{Vis_cmp}
	\end{center}
\end{figure*}
\vspace{-0.5cm}

\begin{table*}[h]
	\centering
	\caption{Value of WPP used to generate different type of warping}
	{\begin{tabular}{|c|c|c|}
			\hline 
			Various types of warping & $P_1$ & $P_2$ \\ \hline 
			Type-I &$0.1, 0.2 , 0.3$ & $0.1, 0.2, 0.3$\\ \hline
			Type-II &$0.7, 0.8, 0.9$& $0.7, 0.8, 0.9$ \\ \hline
			Type-III \& IV &$0.5$& $0.5$ \\ \hline
	\end{tabular}}{}
	\label{var_wpp}
\end{table*}
\vspace{-0.5cm}

\begin{table*}[h]
	\centering
	\caption{Value of WCP used to generate different type of warping}
	{\begin{tabular}{|c|c|c|}
			\hline 
			Types of warping & $P_3^{*}, P_6 ^{*}$ & $P_7^{*}, P_{10} ^{*}$\\ \hline 
			Type- I \& II \& III & $(0.04\times D), \ldots , (0.06\times D)$ & $-(0.04\times D), \ldots ,-(0.06\times D)$\\ \hline
			Type-IV &$-(0.04\times D), \ldots, -(0.06\times D)$ &$(0.04\times D), \ldots, (0.06\times D)$ \\ \hline
			
	\end{tabular}}{}\\
	{$^{*}$ The step size for $P_3, P_6, P_7, P_{10}$ is $(0.005 \times D)$ where $D$ is the length of the diagonal of the image.}
	\label{var_wcp}
\end{table*}
\vspace{-0.7cm}

\section{ Experimental Results and Evalution}
The method is implemented and tested in a PC (Intel(R) Core (TM) i7-6700 3.4 GHz CPU, running Ubuntu 16.04 platform). No specialised hardware is needed to generate the warping factors of each pixel. We have used `HP LaserJet M1005 MFP Multi-function Printer' to scan a document. In our experiment, we have generated images having resolution $300$ dpi and $600$ dpi.  Here, four different types of warping are considered: Type-I. Book having undistorted part at left (as shown in Fig. \ref{Vis_cmp}(a, b)); Type-II. Book having undistorted part at right (as shown in Fig. \ref{Vis_cmp}(c, d)); Type-III. Document pasted on Lamp-post (as shown in Fig. \ref{Vis_cmp}(e, f)); Type-IV. Document attached only at top-middle on a notice board (as shown in Fig. \ref{Vis_cmp}(g, h)).  The undistorted part at left suggests that the perpendicular drawn from optical centre to the document surface, hits the document surface at left side of the middle of the document. The different values of parameters for creating different images are shown in Table \ref{var_wpp} and \ref{var_wcp}.

To visualize the performance of the proposed synthetic warped image generation technique, we have captured some images using mobile camera and the images have different variety of warping. These images are binarized  and the border noises are removed. Many techniques are proposed to binarise a document image \cite{Pratikakis2017}, \cite{Meng_bin_2017}, \cite{Su_bin_2013} and remove border noises from document images \cite{Dey2017}, \cite{Shafait2009}, \cite{Shafait2008}, \cite{Dey2012}, \cite{Bukhari2012_bn} recently.  Here we have used a recent simple and robust binarisation technique proposed by Su et. al. \cite{Su_bin_2013}. Among all the border noise removal techniques most of the approaches works good enough for document images having a flat surface. But method proposed by Bukhari et. al. \cite{Bukhari2012_bn} is specially designed for camera captured warped document image. So, we have used this technique to remove the border noises. The same documents are scanned using flat-bed scanner and binarized. These scanned images are used as the inputs to the proposed warped image generation method and try to generate the warped images which are look like camera captured images.
Fig.~\ref{Vis_cmp} shows a set of pre-processed camera captured images and the corresponding synthetic images generated by the proposed method. It is evident from Fig.~\ref{Vis_cmp} that synthetic warped images is almost similar to their corresponding camera captured images.

\begin{figure}
	\begin{center}
		$\begin{array}{@{\hspace{10pt}}c}
		\fbox{\includegraphics[height=0.9in]{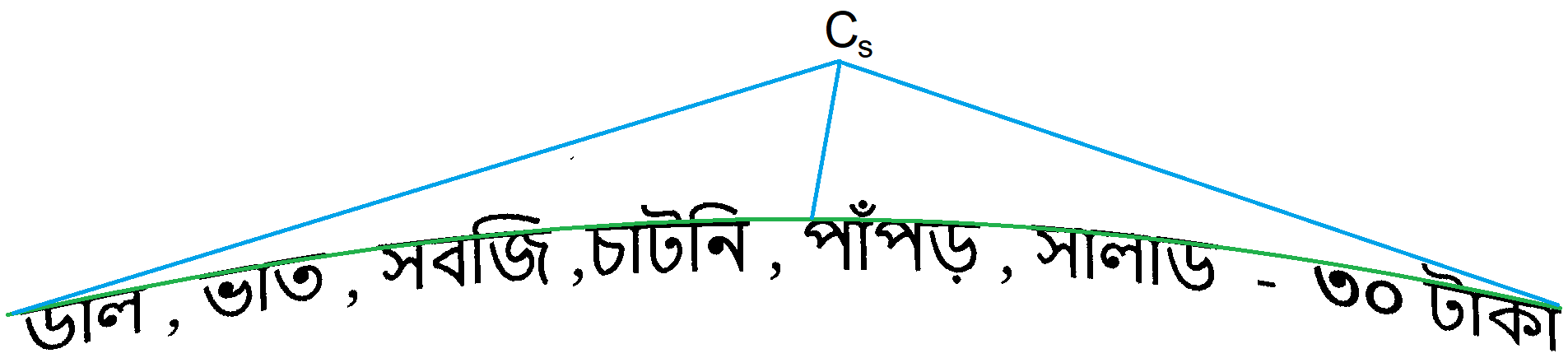}}
		\end{array} $ 
		\caption{Calculation of curvature} 
		\label{cal_crv}
	\end{center}
\end{figure}

To measure the performance of the proposed method, we calculate the curvatures of the `headline' of text lines present in camera captured warped image and the corresponding synthetic warped image.  Here, the curvature is calculated using three points and they are two end points of the `headline' and the point on the `headline' from where the slope changes its sign. Let  $C_r$ and $C_s$ be  the curvature of the `matra/headline' of text lines present in the  real image and synthetic image respectively. An example of a 'headline' for a particular text line is shown in Fig. \ref{cal_crv} and the 'headline is obtained using the method proposed in~\cite{Garai2017}. For each image (real and synthetic), we have considered four text lines to evaluate the performance of the proposed method. Two text lines from the top and two from the bottom of the image. The length of  the each considered text line is greater than $80\%$ of the longest text line present in each image.

The root mean squired error (RMSE), $R_c$ is calculated using the formula: 
$ R_c = \sqrt{\frac{\sum_{i=1}^{N} [C_r(i) - C_s(i)]^2}{N}}$. Here, $N$ is the number of text lines under consideration. We have considered $10$ images from each of four types of warping (Total $40$ images). The average of all the values of $R_c$ is $3.78 \times 10^{-5}$.

\section{Conclusion}
The approach of generating synthetic images proposed by us not only helps in training neural networks but also to measure the performance dewarping approaches. The proposed method can be used to generated different types of warped images. 
The dewarping of non-text part of a document like figures or tables can also be evaluated along with the text part by the proposed technique. 

\begin{spacing}{0.05}
\bibliographystyle{splncs04}
\bibliography{ref}
\end{spacing}

\end{document}